\documentclass{article}

\usepackage{microtype}
\usepackage{graphicx}
\usepackage{subfigure}
\usepackage{amsfonts}
\usepackage{enumitem}
\usepackage{booktabs} % for professional tables

\usepackage{hyperref}

\usepackage{amsmath}
\DeclareMathOperator*{\argmax}{argmax}

\usepackage[accepted]{icml2021}

\icmltitlerunning{On Explainability of Graph Neural Networks via Subgraph Explorations}

\begin{document}

\twocolumn[
\icmltitle{On Explainability of Graph Neural Networks via Subgraph Explorations}



\icmlsetsymbol{equal}{*}

\begin{icmlauthorlist}
\icmlauthor{Hao Yuan}{tamu}
\icmlauthor{Haiyang Yu}{tamu}
\icmlauthor{Jie Wang}{ustc}
\icmlauthor{Kang Li}{sichuan}
\icmlauthor{Shuiwang Ji}{tamu}
\end{icmlauthorlist}

\icmlaffiliation{tamu}{Department of Computer Science \& Engineering, Texas A\&M University, TX, USA}
\icmlaffiliation{ustc}{Department of Electronic Engineering and Information Science, University of Science and Technology of China, Hefei, China}
\icmlaffiliation{sichuan}{West China Biomedical Big Data Center,  West China Hospital, Chengdu, China}

\icmlcorrespondingauthor{Shuiwang Ji}{sji@tamu.edu}

\icmlkeywords{Explainability, Graph Nerual Networks}

\vskip 0.3in
]



\printAffiliationsAndNotice{}  % leave blank if no need to mention equal contribution

\begin{abstract}
We consider the problem of explaining the predictions of graph neural networks (GNNs), which otherwise are considered as black boxes.
Existing methods invariably focus on explaining the importance of graph nodes or edges but ignore the substructures of graphs, which are more intuitive and human-intelligible. In this work, we propose a novel method, known as SubgraphX, to explain GNNs by identifying important subgraphs. Given a trained GNN model and an input graph, our SubgraphX explains its predictions by efficiently exploring different subgraphs with Monte Carlo tree search. To make the tree search more effective, we propose to use Shapley values as a measure of subgraph importance, which can also capture the interactions among different subgraphs. To expedite computations, we propose efficient approximation schemes to compute Shapley values for graph data. {Our work represents the first attempt to explain GNNs via identifying subgraphs explicitly and directly.} Experimental results show that our SubgraphX 
achieves significantly improved explanations, while keeping computations at a reasonable level. 
\end{abstract}

\section{Introduction} \label{intro}
Graph neural networks have drawn significant attention recently due to their promising performance on various graph tasks, including graph classification, node classification, link prediction, and graph generation. Different techniques have been proposed to improve the performance of deep graph models, such as graph convolution~\cite{kipf2016semi, gilmer2017neural, Gao:KDD18, wang2020advanced,yuan2021node2seq}, graph attention~\cite{velivckovic2017graph, wang2019heterogeneous}, and graph pooling~\cite{Yuan2020StructPool,gao2019graph, zhang2018end}. However, these models are still treated as black boxes, and their predictions lack explanations. Without understanding and reasoning the relationships behind the predictions, these models cannot be understood and fully trusted, which prevents their applications in critical areas. This raises the need of investigating the explainability of deep graph models. 

Recently, extensive efforts have been made to study explanation techniques for deep models on images and text~\cite{simonyan2013deep, yuan2019Interpreting, smilkov2017smoothgrad, yuan2020interpreting, yang2019xfake, du2018towards}. These methods can explain both general network behaviors and input-specific predictions via different strategies.
However, the explainability of GNNs is still less explored. Unlike images and texts, graphs are not grid-like data and contain important structural information. Thus, methods for images and texts cannot be applied directly. While several recent studies have developed GNN explanation methods, such as GNNExplainer~\cite{ying2019gnnexplainer}, PGExplainer~\cite{luoparameterized}, and PGM-Explainer~\cite{vu2020pgm}, they invariably focus on explainability at node, edge, or node feature levels, { and only consider subgraphs indirectly via regularization terms.}
We argue that subgraph-level explanations are more intuitive and useful, since subgraphs can be simple building blocks of complex graphs and are highly related to the functionalities of graphs~\cite{alon2007network,milo2002network}. 

In this work, we propose the SubgraphX, a novel GNN explanation method that can identify important subgraphs to explain GNN predictions. Specifically, we propose to employ the Monte Carlo tree search algorithm~\cite{silver2017mastering} to efficiently explore different
subgraphs for a given input graph. Since the information aggregation procedures in GNNs can be interpreted as interactions among different graph structures, we propose to employ Shapley values~\cite{kuhn1953contributions} to measure the importance of subgraphs by capturing such interactions. Furthermore, we propose efficient approximation schemes to Shapley values by considering interactions only within the information aggregation range.
Altogether, our work represents the first attempt to explain GNNs via identifying subgraphs explicitly.
We conduct both qualitative and quantitative experiments to evaluate the effectiveness and efficiency of our SubgraphX.
Experimental results show that our proposed SubgraphX can provide better explanations for a variety of GNN models. In addition, our method has a reasonable computational cost given its superior performance.

\section{Related Work}

\subsection{Graph Neural Networks}
Graph neural networks have demonstrated their effectiveness on different graph tasks. Several approaches are proposed to learn representations for nodes and graphs, such as GCNs~\cite{kipf2016semi}, GATs~\cite{velivckovic2017graph}, and GINs~\cite{xu2018powerful}, etc. These methods generally follow an information aggregation scheme that the features of a target node are obtained by aggregating and combining the features from its neighboring nodes. Here we use GCNs as an example to illustrate such information aggregation procedures. 
Formally, a graph $\mathcal{G}$ with $m$ nodes can be represented by an adjacency matrix $A \in \{0,1\}^{m\times m}$ and a feature matrix $X \in \mathbb{R}^{m\times d}$ assuming that each node is associated with a $d$-dimensional feature vector. Then the aggregation operation in GCNs can be mathematically written as  
$X_{i+1} = \sigma(D^{-\frac{1}{2}}\hat{A}D^{-\frac{1}{2}}X_{i}W_{i})$,
where $X_{i}$ denotes the output feature matrix of $i-$th GCN layer and $X_0$ is set to $X_0=X$. The node features are  transformed from $X_{i} \in \mathbb{R}^{m\times c_{i}}$ to $X_{i+1} \in \mathbb{R}^{m\times c_{i+1}}$. 
Note that $\hat{A}=A+I$ is employed to add self-loops and $D$ is a diagonal node degree matrix to perform normalization on $\hat{A}$. In addition, 
$W_{i}\in \mathbb{R}^{c_{i}\times c_{i+1}}$ is a
learnable weight matrix to perform linear transformations on features and $\sigma(\cdot)$ is the non-linear activation function.

\subsection{Explainability in Graph Neural Networks}
Even though explaining GNNs is crucial to understand and trust deep graph models, the explainability of GNNs is still less studied, compared with the image and text domains. 
Recently, several methods are proposed specifically to explain deep graph models. 
These methods mainly focus on explaining GNNs by identifying important nodes, edges, node features. 
However, none of them can provide input-dependent subgraph-level explanations, which is important for understanding graph models. 
Following a recent survey work~\cite{yuan2020explainability},  we categorize these methods into several classes; those are, gradients/features-based methods, decomposition methods, surrogate methods, generation-based methods, and perturbation-based methods.

First, several methods employ gradient values or feature values to study the importance of input graph nodes, edges, or node features~\cite{baldassarre2019explainability, pope2019explainability}. These methods generally extend existing image explanation techniques to the graph domain, such as SA~\cite{zeiler2014visualizing}, CAM~\cite{zhou2016learning}, and Guided BP~\cite{springenberg2014striving}. While these methods are simple and efficient, they cannot incorporate the special properties of graph data. Meanwhile, decomposition methods, such as LRP~\cite{schwarzenberg2019layerwise}, Excitation BP~\cite{pope2019explainability}, and GNN-LRP~\cite{schnake2020xai}, explain GNNs by decomposing the original model predictions into several terms and associating these terms with graph nodes or edges. These methods generally follow a backpropagation manner to decompose predictions layer by layer until input space.  In addition, existing methods~\cite{huang2020graphlime, vu2020pgm} employ a simple and interpretable model as the surrogate method to capture local relationships of deep graph models around the input data. Then the explanations of the surrogate method are treated as the explanations of the original predictions. Furthermore,  recent work XGNN~\cite{xgnn} proposes to study general and high-level explanations of GNNs by generating graph patterns to maximize a certain prediction. 

In addition, a popular direction to explain GNNs is known as the perturbation-based method. It monitors the changes in the predictions by perturbing different input features and identifies the features affecting predictions the most. For example,  GNNExplainer~\cite{ying2019gnnexplainer} optimizes soft masks for edges and node features to maximize the mutual information between the original predictions and new predictions. Then the optimized masks can identify important edges and features. Meanwhile,  PGExplainer~\cite{luoparameterized} learns a parameterized model to predict whether an edge is important, which is trained using all edges in the dataset. It employs the reparameterization trick~\cite{jang2016categorical} to obtain approximated discrete masks instead of soft masks. In addition, GraphMask~\cite{schlichtkrull2020interpreting} follows a similar idea as PGExplainer that train a classifier to predict if an edge can be dropped without affecting model predictions. However, it studies the edges in every GNN layer while PGExplainer only focuses on the input space. 

\begin{figure*}[ht!]
    \centering
    \includegraphics[width=1.76\columnwidth]{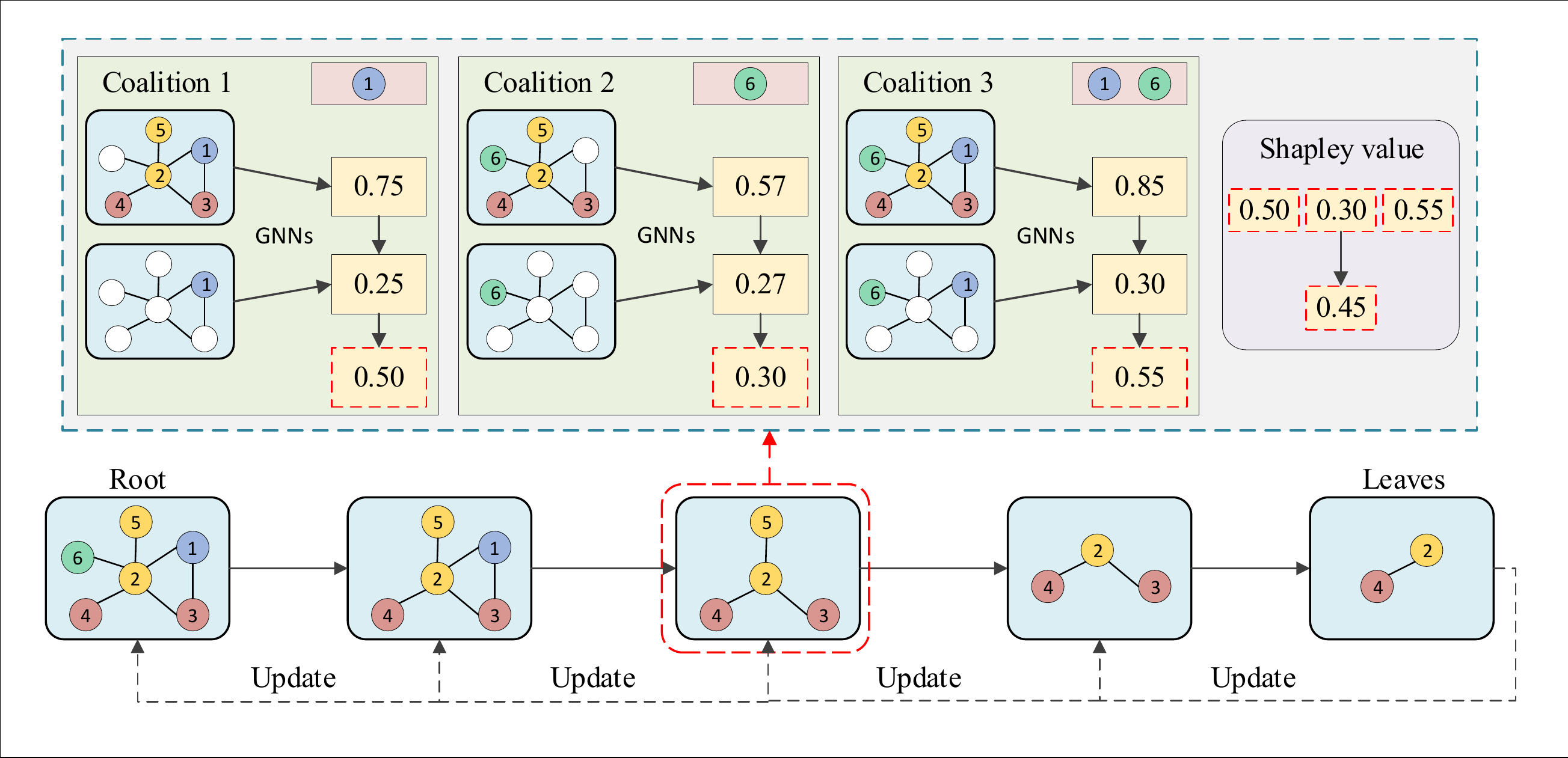}
    %\vspace{-0.5cm}
    \caption{An illustration of our proposed SubgraphX. The bottom shows one selected path from the root to leaves in the search tree, which corresponds to one iteration of MCTS. For each node, its subgraph is evaluated by computing the Shapley value via Monte-Carlo sampling.
    In this example, we show the computation of Shapley value for the middle node (shown in red dashed box) where three coalitions are sampled to compute the marginal contributions. 
    Note that nodes that are not selected are ignored for simplicity.}
    \label{fig:xsub}
\end{figure*}

\section{The Proposed SubgraphX}

{While most current methods for GNN explanations are invariably based on directly identifying important nodes or edges, we argue that directly studying important subgraphs is more natural and may lead to better explainability.} In this work, we propose a novel approach, known as SubgraphX, to explain GNNs by exploring and identifying important subgraphs.

\subsection{From Node and Edge to Subgraph Explanations}

Unlike images and texts, graph data contain important structural information, which is highly related to the properties of graphs. For example, network motifs, which can be considered as graph substructures, are simple building blocks of complex networks and may determine the functionalities of graphs in many domains, such as biochemistry, ecology, neurobiology, and engineering~\cite{alon2007network,milo2002network,shen2002network,alon2019introduction}. Hence, investigating graph substructures is a crucial step towards the reverse engineering and understanding of the underlying mechanisms of GNNs. In addition, subgraphs are more intuitive and human-intelligible~\cite{yuan2020explainability}.

While different methods are proposed to explain GNNs, none of them can directly provide subgraph-level explanations for individual input examples. The XGNN can obtain graph patterns to explain GNNs but its explanations are not input-dependent and less precise.  The other methods, such as GNNExplainer and PGExplainer, may obtain subgraph-level explanations by combining nodes or edges to form subgraphs in a post-processing manner. However, the important nodes or edges in their explanations are not guaranteed to be connected. 
Meanwhile, since GNNs are very complex, node/edge importance cannot be directly converted to subgraph importance. 
Furthermore, these methods ignore the interactions among different nodes and edges, which may contain important information. Hence, in this work, we propose a novel method, known as SubgraphX, to directly study the subgraphs to provide explanations. The explanations of our SubgraphX are connected subgraphs, which are more human-intelligible. In addition, by incorporating Shapley values, our method can capture the interactions among different graph structures when providing explanations.

\subsection{Explaining GNNs with Subgraphs}
We first present a formal problem formulation. Let $f(\cdot)$ denote the trained GNNs to be explained. Without loss of generality,
we introduce our proposed SubgraphX by considering $f(\cdot)$ as a graph classification model. Given an input graph $\mathcal{G}$, its predicted class is represented as ${y}$.
%and the corresponding predicted probability is $p=f(\mathcal{G})_y$. 
The goal of our explanation task is to find the most important subgraph for the prediction ${y}$. Since disconnected nodes are hard to understand, we only consider connected subgraphs to enable the explanations to be more human-intelligible. Then the set of connected subgraphs of $\mathcal{G}$ is denoted as $\{\mathcal{G}_1, \cdots,\mathcal{G}_i, \cdots, \mathcal{G}_n \}$ where $n$ is the number of different connected subgraphs in $\mathcal{G}$. 
The explanation of prediction ${y}$ for input graph $\mathcal{G}$ can then be defined as 
\begin{equation}\label{eq1}
\mathcal{G}^*= \argmax_{|\mathcal{G}_i|\leq N_{\min}} \mbox{Score}(f(\cdot), \mathcal{G}, \mathcal{G}_i ), 
\end{equation}
where $\mbox{Score}(\cdot,\cdot,\cdot)$ is a scoring function for evaluating the importance of a subgraph given the trained GNNs and the input graph. We use $N_{\min}$ as an upper bound on the size of subgraphs so that the obtained explanations are succinct enough. 
A straightforward way to obtain $\mathcal{G}^*$ is to enumerate all possible $\mathcal{G}_i$ and select the most important one as the explanation. However, such a brute-force method is intractable when the graph is complex and large-scale. Hence, in this work, we propose to incorporate search algorithms to explore subgraphs efficiently.  
Specifically, we propose to employ Monte Carlo Tree Search (MCTS)~\cite{silver2017mastering, jin2020multi} as the search algorithm. 
In addition, since the information aggregation procedures in GNNs can be understood as interactions between different graph structures, we propose to employ the Shapley value~\cite{kuhn1953contributions} as the scoring function to measure the importance of different subgraphs by considering such interactions. We illustrate our proposed SubgraphX in Figure~\ref{fig:xsub}. 
After searching, the subgraph with the highest score is considered as the explanation of the prediction $y$ for input graph $\mathcal{G}$. Note that our proposed SubgraphX can be easily extended to use other search algorithms and scoring functions. 

\subsection{Subgraph Exploration via MCTS}
In our proposed SubgraphX, we employ the MCTS as the search algorithm to guide our subgraph explorations. We build a search tree in which the root is associated with the input graph and each of other nodes corresponds to a connected subgraph. Each edge in our search tree denotes that the graph associated with a child node can be obtained by performing node-pruning from the graph associated with its parent node.
Formally, we define a node in this search tree as $\mathcal{N}_i$, and $\mathcal{N}_0$ denotes the root node. The edges in the search tree represent the pruning actions $a$. Note that 
each node may have many pruning actions, and these actions can be defined based on the dataset at hand or domain knowledge. Then the MCTS algorithm records the statistics of visiting counts and rewards to guide the exploration and reduce the search space. 
Specifically, for the node and pruning action pair $(\mathcal{N}_i, a_j)$, we assume that the subgraph $\mathcal{G}_j$ is obtained by action $a_j$ from $\mathcal{G}_i$. Then the 
MCTS algorithm records four variables for $(\mathcal{N}_i, a_j)$, which are defined as:
\begin{itemize}[noitemsep, topsep=0pt,leftmargin=*]
    \item $C(\mathcal{N}_i, a_j)$ denotes the number of counts for selecting action $a_j$ for node $\mathcal{N}_i$.
    \item $W(\mathcal{N}_i, a_j)$ is the total reward for all $(\mathcal{N}_i, a_j)$ visits.
    \item $Q(\mathcal{N}_i, a_j)=W(\mathcal{N}_i, a_j)/C(\mathcal{N}_i, a_j)$ and denotes the averaged reward for multiple visits.
    \item $R(\mathcal{N}_i, a_j)$ is the immediate reward for selecting $a_j$ on $\mathcal{N}_i$, which is used to measure the importance of subgraph $\mathcal{G}_j$. We propose to use $R(\mathcal{N}_i, a_j)= \mbox{Score}(f(\cdot), \mathcal{G}, \mathcal{G}_j)$. 
\end{itemize}

In each iteration, the MCTS selects a path starting from the root $\mathcal{N}_0$ to a leaf node $\mathcal{N}_\ell$. Note that the leaf nodes can be defined based on the numbers of nodes in subgraphs such that $|\mathcal{N}_\ell| \leq N_{\min}$.
Formally, the action selection criteria of node $\mathcal{N}_i$ are defined as 
\begin{align}
a^* &= \argmax_{a_j} Q(\mathcal{N}_i, a_j) +  U(\mathcal{N}_i, a_j),\label{eq2} \\
U(\mathcal{N}_i, a_j) &=  \lambda R(\mathcal{N}_i, a_j)\frac{\sqrt{\sum_{k} C(\mathcal{N}_i, a_k)}}{1+ C(\mathcal{N}_i, a_j)},\label{eq3}
\end{align}
where $\lambda$ is a hyperparameter to control the trade-off between exploration and exploitation. In addition, $\sum_{k} C(\mathcal{N}_i, a_k)$ denotes the total visiting counts for all possible actions of  node $\mathcal{N}_i$. Then the subgraph in the leaf node $\mathcal{N}_\ell$ is evaluated and the importance score is denoted as $\mbox{Score}(f(\cdot), \mathcal{G}, \mathcal{G}_\ell)$. Finally, all node and action pairs selected in this path are updated as 
\begin{align}
C(\mathcal{N}_i, a_j) &= C(\mathcal{N}_i, a_j) + 1 ,\label{eq4} \\
W(\mathcal{N}_i, a_j) &=  W(\mathcal{N}_i, a_j) + \mbox{Score}(f(\cdot), \mathcal{G}, \mathcal{G}_\ell).\label{eq5}
\end{align}
After searching for several iterations, we select the subgraph with the highest score from the leaves as the explanation. Note that in early iterations, the MCTS tends to select child nodes with low visit counts in order to explore different possible pruning actions. In later iterations, the MCTS tends
to select child nodes that yield higher rewards, \emph{i.e.}, more important subgraphs.

\subsection{A Game-Theoretical Scoring Function}
In our proposed SubgraphX, both the MCTS rewards and the explanation selection are highly depending on the scoring function $\mbox{Score}(\cdot,\cdot,\cdot)$. It is crucial to properly measure the importance of different subgraphs. One possible solution is to directly feed the subgraphs to the trained GNNs $f(\cdot)$ and use the predicted scores as the importance scores. However, it cannot capture the interactions between different graph structures, thus affecting the explanation results. Hence, in this work, we propose to adopt the Shapley values~\cite{kuhn1953contributions, lundberg2017unified, chen2018shapley} as the scoring function. The Shapley value is a solution concept from the cooperative game theory for fairly assigning a total game gain to different game players. To apply it to graph model explanation tasks, we use the GNN prediction as the game gain and different graph structures as players.   

Formally, given the input graph $\mathcal{G}$ with $m$ nodes and the trained GNN $f(\cdot)$, we study the Shapley value for a target subgraph $\mathcal{G}_i$ with $k$ nodes. Let $V = \{v_1, \cdots, v_i, \cdots, v_m \}$ denote all nodes in $\mathcal{G}$ and we assume that the nodes in $\mathcal{G}_i$ are $\{v_1, \cdots, v_k\}$ while the other nodes $\{v_{k+1}, \cdots, v_m\}$ belong to $\mathcal{G}\setminus \mathcal{G}_i$. Then the set of players is defined as $P= \{\mathcal{G}_i, v_{k+1}, \cdots, v_m\}$, where we consider the whole subgraph $\mathcal{G}_i$ as one player.  Finally, the Shapley value of the player $\mathcal{G}_i$ can be  computed as
\begin{align}
\hspace{-0.3cm}\phi(\mathcal{G}_i)&=\sum_{S\subseteq P\setminus\{\mathcal{G}_i\}}\frac{|S|!\left(|P|-|S|-1\right)!}{|P|!}m(S,{G}_i),\label{eq:6}\\
\hspace{-0.3cm}m(S,\mathcal{G}_i) &= f\left(S\cup\{\mathcal{G}_i\}\right)-f(S),\label{eq:7}
\end{align}
where $S$ is the possible coalition
set of players. Note that $m(S,\mathcal{G}_i)$ represents  the marginalized contribution of player $\mathcal{G}_i$ given the coalition set $S$. It can be computed by the difference of predictions between incorporating $\mathcal{G}_i$ with and without the coalition set $S$. The obtained Shapley value $\phi(\mathcal{G}_i)$ considers all different coalitions to capture the interactions. It is the only solution that satisfies four desirable axioms, including efficiency, symmetry, linearity, and dummy axiom~\cite{lundberg2017unified}, which can guarantee the correctness
and fairness of the explanations. However, computing Shapley values using Eqs. (\ref{eq:6}) and (\ref{eq:7}) is time-consuming as it enumerates all possible coalitions, especially for large-scale and complex graphs. Hence, in this work, we propose to incorporate the GNN architecture information $f(\cdot)$ to efficiently approximate Shapley values.

\subsection{Graph Inspired Efficient Computations}

In graph neural networks, the new features of a target node are obtained by aggregating information from a limited neighboring region. Assuming there are $L$ layers of GNN in the graph model $f(\cdot)$, then only the neighboring nodes within $L$-hops are used for information aggregation. Note that the information aggregation schema can be considered as interactions between different graph structures. Hence, the subgraph $\mathcal{G}_i$ mostly interacts with the neighbors within $L$-hops. Based on such observations, we propose to compute the Shapley value of $\mathcal{G}_i$ by only considering its $L$-hop neighboring nodes. Specifically, assuming there are $r$ ($r\leq m-k$) nodes within $L$-hop neighboring of subgraph $\mathcal{G}_i$, we denote these nodes as $\{v_{k+1}, \cdots, v_r\}$. Then the new set of players we need to consider is represented as 
$P'= \{\mathcal{G}_i, v_{k+1}, \cdots, v_r\}$. By incorporating  $P'$, the Shapley value of $\mathcal{G}_i$ can be defined as 
\begin{equation}\label{eq:9}
\phi(\mathcal{G}_i)=\sum_{S\subseteq P'\setminus\{\mathcal{G}_i\}}\frac{|S|!\left(|P'|-|S|-1\right)!}{|P'|!}m(S,{G}_i).
\end{equation}

\begin{algorithm}[tb]
   \caption{The algorithm of our proposed SubgraphX.}
   \label{alg:algo1}
\begin{algorithmic}
   \STATE {\bfseries Input:} GNN model $f(\cdot)$, input graph $\mathcal{G}$, MCTS iteration number $M$, the leaf threshold node number $N_{\min}$, $h(\mathcal{N}_i)$ denotes the associated subgraph of tree node $\mathcal{N}_i$. 
    \STATE {\bfseries Initialization:} for each $(\mathcal{N}_i, a_j)$ pair , initialize its $C$, $W$, $Q$, and $R$ variables as 0. The root of search tree is $\mathcal{N}_0$ associated with graph $\mathcal{G}$. The leaf set is set to  $S_\ell = \{\}$. 
   \FOR{$i=1$ {\bfseries to} $M$}
   \STATE $curNode = \mathcal{N}_0$, $curPath = [\mathcal{N}_0]$
   \WHILE{ $h(curNode)$ has more node than $N_{\min}$}
   \FOR{all possible pruning actions of $h(curNode)$ }
   \STATE Obtain child node $\mathcal{N}_j$ and its subgraph $\mathcal{G}_j$.
   \STATE  Compute $R(curNode, a_j)= \mbox{Score}(f(\cdot), \mathcal{G}, \mathcal{G}_j))$ with Algorithm~\ref{alg:algo2}.
   \ENDFOR
   \STATE Select the child $\mathcal{N}_{next}$ following Eq.(\ref{eq2}, \ref{eq3}).
   \STATE $curNode = \mathcal{N}_{next}$, $curPath = curPath + \mathcal{N}_{next}$.
   \ENDWHILE
   \STATE $S_\ell= S_\ell \cup \{curNode\}$
   \STATE Update nodes in $curPath$ following Eq.(\ref{eq4}, \ref{eq5}).
   \ENDFOR
   \STATE Select subgraph with the highest score from $S_\ell$.
\end{algorithmic}
\end{algorithm}
However, since graph data are complex that different nodes have variable numbers of neighbors, then $P'$ may still contain a large number of players, thus affecting the efficiency of computation. Hence, in our SubgraphX, we further incorporate the Monte-Carlo sampling~\cite{vstrumbelj2014explaining} to compute $\phi(\mathcal{G}_i)$. 
Specifically, for sampling step $i$, we sample a coalition set $S_i$ from the player set $P'\setminus\{\mathcal{G}_i\}$ and compute its marginalized contribution
$m(S_i,\mathcal{G}_i)$. Then the averaged contribution score for multiple sampling steps is regarded as the approximation of $\phi(\mathcal{G}_i)$. Formally, it can be mathematically written as
\begin{equation}\label{eq:10}
\phi(\mathcal{G}_i)= \frac{1}{T}\sum_{t=1}^T (f\left(S_i\cup\{\mathcal{G}_i\}\right)-f(S_i)),
\end{equation}
where $T$ is the total sampling steps. In addition, to compute the marginalized contribution, we follow a zero-padding strategy. Specifically, to compute $f\left(S_i\cup\{\mathcal{G}_i\}\right)$, we consider the nodes $V\setminus \left(S_i\cup\{\mathcal{G}_i\}\right)$ which are not belonging to the coalition or the subgraph and set their node features to all zeros. Then we feed the new graph to the GNNs $f(\cdot)$ and use the predicted probability as $f\left(S_i\cup\{\mathcal{G}_i\}\right)$. Similarly, we can compute $f(S_i)$ by setting nodes $V\setminus S_i$ with zero features and feeding to the GNNs. It is noteworthy that we only perturb the node features instead of removing the nodes from the input graph because graphs are very sensitive to structural changes~\cite{schlichtkrull2021interpreting}. Finally, we conclude the computation steps of our proposed SubgraphX in Algorithm~\ref{alg:algo1} and~\ref{alg:algo2}. {Note that $N_{min}$ determines the stop condition in MCTS, and we can select subgraphs with specific sizes from internal nodes of the search tree as needed.}

\begin{algorithm}[tb]
   \caption{The algorithm of subgraph Shapley value.}
   \label{alg:algo2}
\begin{algorithmic}
   \STATE {\bfseries Input:} GNN model $f(\cdot)$ with $L$ layers, input graph $\mathcal{G}$ with nodes $V=\{v_1,\dots,v_m\}$, subgraph $\mathcal{G}_i$ with $k$ nodes $\{v_1,\dots,v_k\}$, Monte-Carlo sampling steps $T$.
   \STATE {\bfseries Initialization:} Obtain the $L$-hop neighboring nodes of $\mathcal{G}_i$, denoted as $\{v_{k+1}, \cdots, v_r\}$. Then the set of players is $P'= \{\mathcal{G}_i, v_{k+1}, \cdots, v_r\}$.
   \FOR{$i=1$ {\bfseries to} $T$}
   \STATE Sampling a coalition set $S_i$ from $P'\setminus\{\mathcal{G}_i\}$. 
   \STATE Set nodes from $V\setminus \left(S_i\cup\{\mathcal{G}_i\}\right)$ with zero features and feed to the GNNs $f(\cdot)$ to obtain $f(S_i\cup\{\mathcal{G}_i\})$.  
   \STATE Set nodes from $V\setminus S_i$ with zero features and feed to the GNNs $f(\cdot)$ to obtain $f(S_i)$. 
   \STATE Then $m(S_i, \mathcal{G}_i)= f(S_i\cup\{\mathcal{G}_i\})- f(S_i)$.
   \ENDFOR
   \STATE {\bfseries Return:} $\mbox{Score}(f(\cdot), \mathcal{G}, \mathcal{G}_i) = \frac{1}{T}\sum_{t=1}^T m(S_i, \mathcal{G}_i)$.
\end{algorithmic}
\end{algorithm}

\subsection{SubgraphX for Generic Graph Tasks}

We have described our proposed SubgraphX using graph classification models as an example. It is noteworthy that our SubgraphX can be easily generalized to explain graph models on other tasks, such as node classification and link prediction. For node classification models, the explanation target is the prediction of a single node $v_i$ given the input graph $\mathcal{G}$. 
\begin{table*}[!ht]
    \caption{Statistics and properties of five datasets.}
    \label{table:data}
    \centering
    \begin{tabular}{@{}lccccc@{}}
        \toprule
        \multicolumn{1}{l}{ }       &    \multicolumn{5}{c}{\textbf{Dataset}}          \\
        \cmidrule(r){2-6}
        & \textsc{MUTAG}     & \textsc{BBBP}& {\textsc{Graph-SST2}} & {\textsc{BA-2Motifs}} & {\textsc{BA-Shape}} \\
        \midrule

        \# of Edges (avg)  & 19.79 & 25.95 & 9.20 & 25.48  & 2055 \\
        \# of Nodes (avg)     & 17.93  & 24.06  & 10.19 & 25.0  & 700   \\
        \# of Graphs     & 188  & 2039 & 70042 & 1000 & 1 \\
        \# of Classes     & 2  & 2 & 2 & 2 & 4    \\
        \bottomrule
    \end{tabular}
\end{table*}
Assuming there are $L$ layers in the GNN models, the prediction of $v_i$ only relies on its $L$-hop computation graph, denoted as $\mathcal{G}_c$. Then instead of searching from the input graph $\mathcal{G}$, our SubgraphX sets $\mathcal{G}_c$ as the corresponding graph of the search tree root $\mathcal{N}_0$. In addition, when computing the marginalized contributions, the zero-padding strategy should exclude the target node $v_i$. Meanwhile, for link prediction tasks, the explanation target is the prediction of a single link $(v_i, v_j)$. Then the root of the search tree corresponds to the $L$-hop 
computation graph of node $v_i$ and $v_j$. Similarly, the zero-padding strategy ignores the $v_i$ and $v_j$ when perturbing node features. Note that our SubgraphX treats the GNNs as black boxes during the explanation stage and only needs to access the inputs and outputs.
Hence, our proposed SubgraphX can be applied to a general family of GNN models, including but not limited to GCNs~\cite{kipf2016semi}, GATs~\cite{velivckovic2017graph}, GINs~\cite{xu2018powerful}, and Line-Graph NNs~\cite{chen2017supervised}.

\section{Experimental Studies} 

\subsection{Datasets and Experimental Settings}
We conduct extensive experiments on different datasets and GNN models to demonstrate the effectiveness of our proposed method. The statistics and properties of the datasets are reported in Table~\ref{table:data}.
We evaluate our SubgraphX with five datasets for both graph classification and node classification tasks, including synthetic data, biological data, and text data. We summarize these datasets as below:
\begin{itemize}[noitemsep, topsep=0pt,leftmargin=*]
\item MUTAG~\cite{debnath1991structure} and BBBP~\cite{wu2018moleculenet} are molecular datasets for graph classification tasks. In these datasets, each graph represents a molecule while nodes are atoms and edges are bonds. The labels are determined by the chemical functionalities of molecules.
\item Graph-SST2~\cite{yuan2020explainability} is sentiment graph dataset for graph classification. It converts text sentences to graphs with Biaffine parser~\cite{gardner2018allennlp} that nodes denote words and edges represent the relationships between words. Note that node embeddings are initialized as the pre-trained BERT word embeddings~\cite{devlin2018bert}. 
Each graph is labeled by its sentiment, which can be positive or negative.  
\item BA-2Motifs is a synthetic graph classification dataset. Each graph contains a based graph generated by \textit{Barabási-Albert} (BA) model, which is connected with a house-like motif or a five-node cycle motif. The graphs are labeled based on the type of motifs. All node embeddings are initialized as vectors containing all 1s. 
\item BA-Shape is a synthetic node classification dataset. Each graph contains a base BA graph and several  house-like five-node motifs. The node labels are determined by the memberships and locations of different nodes. All node embeddings are initialized as vectors containing all 1s.  
\end{itemize}

\begin{figure*}[ht!]
    \centering
    \includegraphics[width=2\columnwidth]{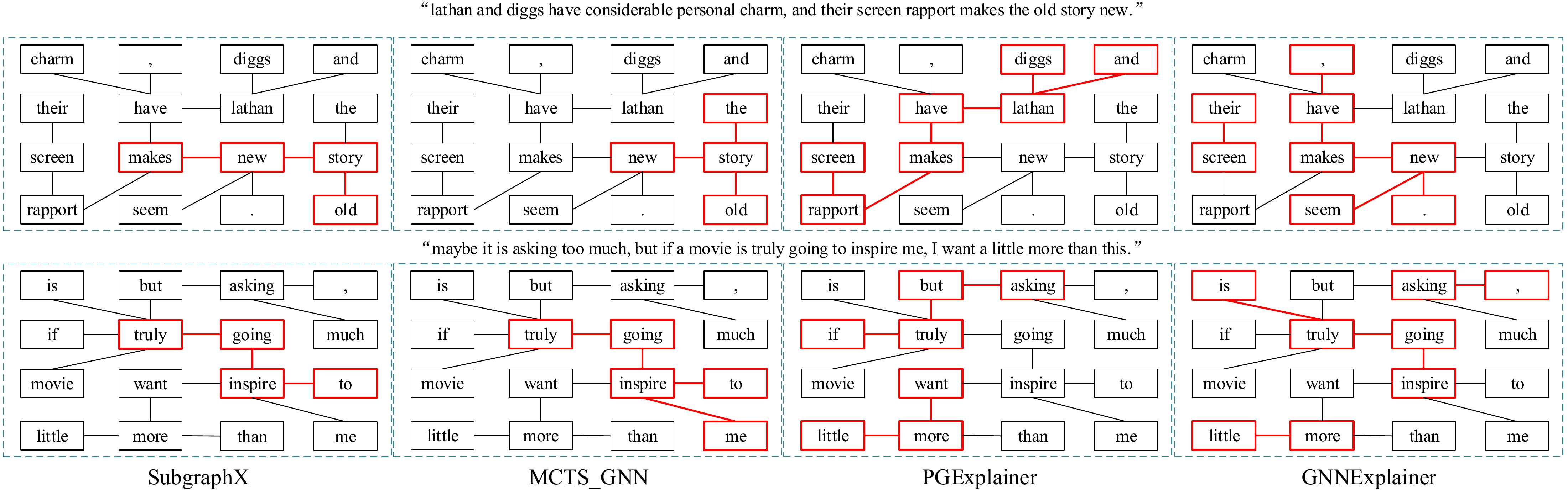}
    %\vspace{-0.4cm}
    \caption{Explanation results on the Graph-SST2 dataset with a GAT graph classifier. The input sentences are shown on the top of explanations. Note that some ``unimportant'' words are ignored for simplicity. 
    The first row shows explanations for a correct prediction and the second row reports the results for an incorrect prediction.}
    \label{fig:vis3}
\end{figure*}

\begin{figure}[t]
    \centering
    \includegraphics[width=\columnwidth]{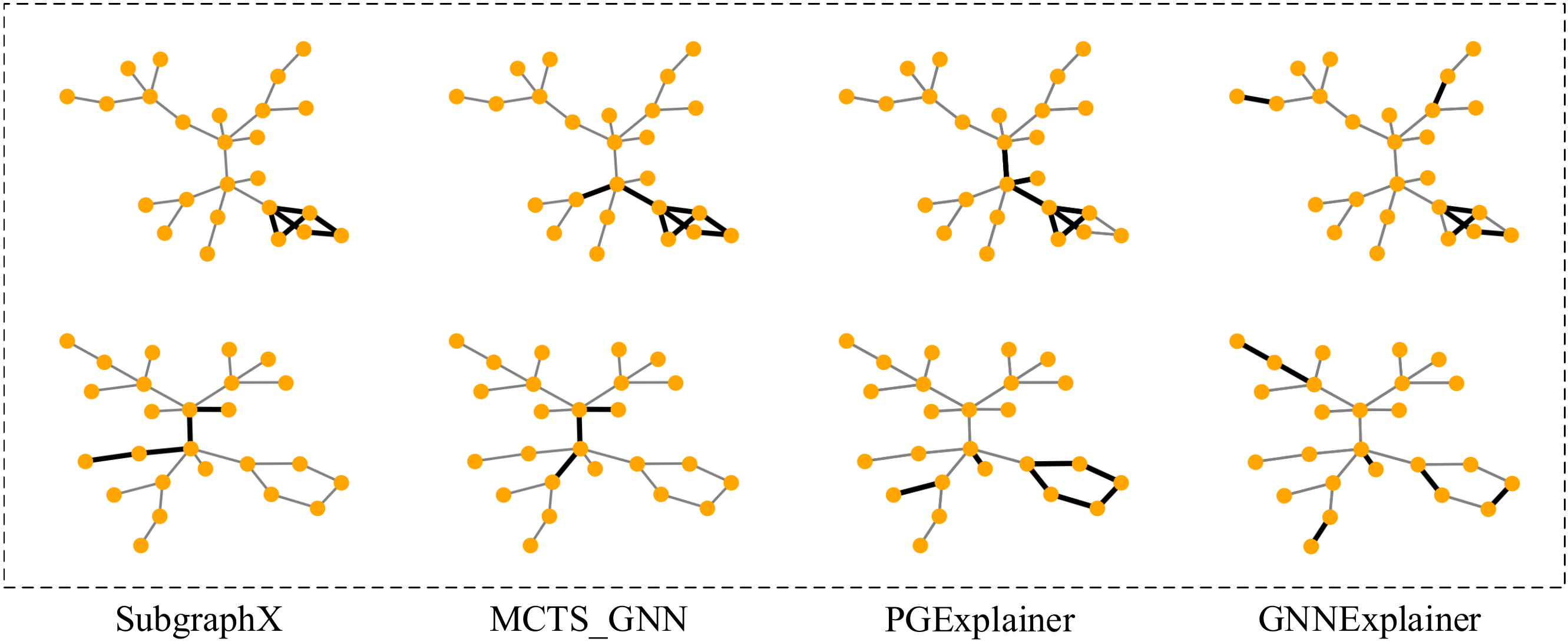}
    %\vspace{-0.7cm}
    \caption{Explanation results on the BA-2Motifs dataset with a GCN graph classifier. The first row shows explanations for a correct prediction and the second row reports the results for an incorrect prediction.}
    \label{fig:vis1}
\end{figure}
\begin{figure}[t]
    \centering
    \includegraphics[width=\columnwidth]{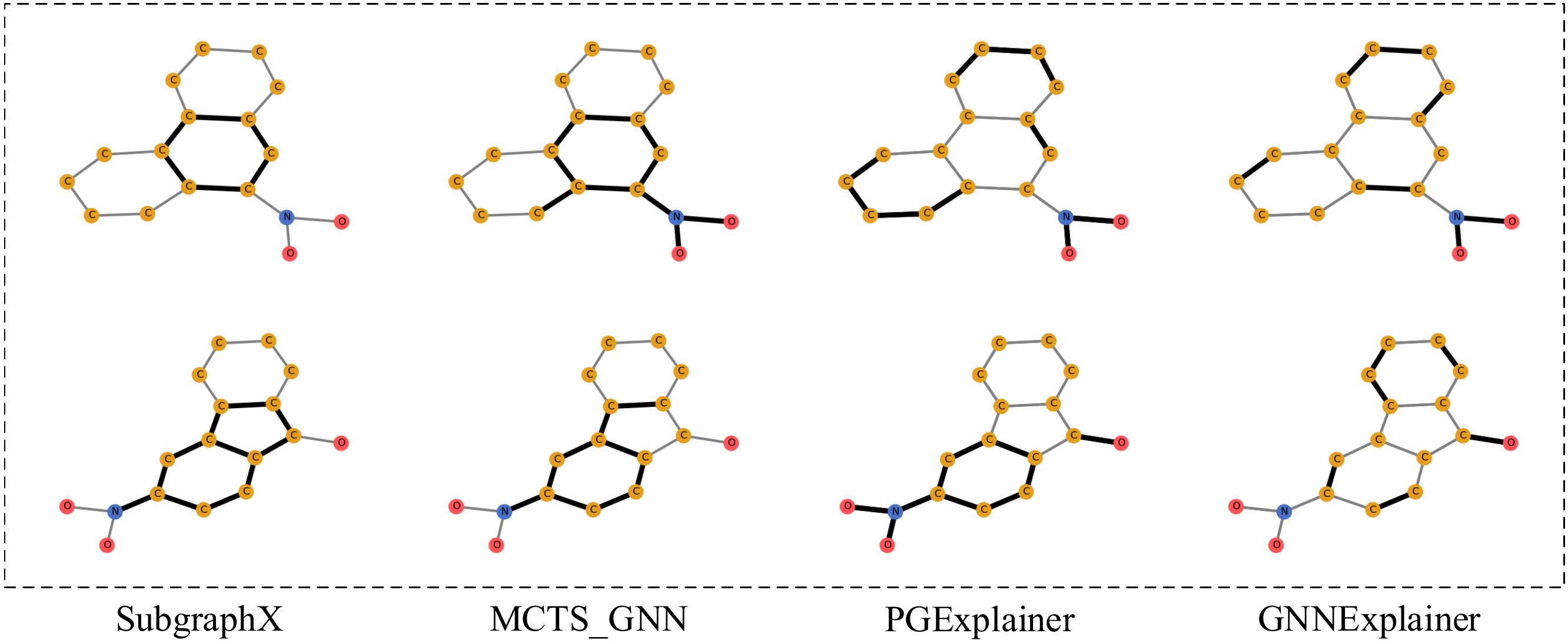}
    %\vspace{-0.7cm}
    \caption{Explanation results on the MUTAG dataset with a GIN graph classifier. We show the explanations for two correct predictions. Here Carbon, Oxygen, and Nitrogen are shown in yellow, red, and blue, respectively.  }
    \label{fig:vis2}
\end{figure}
We explore three variants of GNNs on these datasets, including GCNs, GATs, and GINs. All GNN models used in our experimental studies are trained to obtain reasonable performance. Then we compare our SubgraphX with several baselines, including MCTS\_GNN, GNNExplainer~\cite{ying2019gnnexplainer}, PGExplainer~\cite{luoparameterized}. {Note that GNNExplainer and PGExplainer represent the 
state-of-the-art methods for GNN explanations.}
Here MCTS\_GNN denotes the method using MCTS to explore subgraphs but directly employing the GNN predictions of these subgraphs as the scoring function. 
{We wish to mention that all methods are compared with a fair setting. We use the same number to control the maximum number of nodes in the explanations for all methods.}

We conduct our experiments using one Nvidia V100 GPU on an Intel Xeon Gold 6248 CPU. Our implementations are based on Python 3.7.6, PyTorch 1.6.0, and Torch-geometric 1.6.3. For our proposed SubgraphX and other algorithms with MCTS, the MCTS iteration number $M$ is set to 20. To explore a suitable trade-off between exploration and exploitation, we set the hyperparameter $\lambda$ in Eq.(\ref{eq3}) to 5 for Graph-SST2 (GATs) and BBBP (GCNs) models, and 10 for other models. Since all GNN models contain 3 network layers, we consider 3-hop computational graphs to compute Shapley values for our SubgraphX. For the Monte-Carlo sampling in our SubgraphX, we set the Monte-Carlo sampling steps  $T$ to 100 for all datasets. For MCTS$^\dagger$, we set Monte-Carlo sampling steps to 1000 to obtain good approximations since it samples from all nodes in a graph. 
More details regarding the datasets, trained GNN models, and experimental settings can be found in Supplementary Section~\ref{sup:data}. Our code and data are now publicly available in the DIG library~\cite{liu2021dig}\footnote{\url{https://github.com/divelab/DIG}}.

\subsection{Explanations for Graph Classification Models}
We first visually compare our SubgraphX with the other baselines using graph classification models. The results are reported in Figure~\ref{fig:vis1},~\ref{fig:vis2}, and~\ref{fig:vis3} where important substructures are shown in the bold.

The explanation results of the BA-2Motifs dataset are visualized in Figure~\ref{fig:vis1}. 
We use the GCNs as the graph classifier and report explanations for both correct and incorrect predictions. Since it is a synthetic dataset, we may consider the motifs as reasonable approximations of explanation ground truth. In the first row, the model prediction is correct and our SubgraphX can precisely identify the house-like motif as the most important subgraph. In the second row,
our SubgraphX explains the incorrect prediction that the GNN model cannot capture the five-node cycle motif as the important structure, and hence the prediction is wrong. {For both cases, our SubgraphX can provide better visual explanations since our method can precisely identifies the succinct subgraphs that can reasonably explain the predictions.}
In addition, our explanations are connected subgraphs while PGExplainer and GNNExplainer identify discrete edges.

We also show the explanation results of the MUTAG dataset in Figure~\ref{fig:vis2}. Note that GINs are employed as the graph classification model to be explained. Since the MUTAG dataset is a real-world dataset and there is no ground truth for explanations, we evaluate the explanation results based on chemical domain knowledge. The graphs in MUTAG are labeled based on the mutagenic effects on a bacterium. It is known that carbon rings and $NO_2$ groups tend to be mutagenic~\cite{debnath1991structure}. We study whether the explanations provided by different methods can match the carbon rings and $NO_2$ groups identified in chemistry.
In both examples, the predictions are ``mutagenic'' and our SubgraphX successfully and precisely identifies the carbon rings as important subgraphs. Meanwhile, the MCTS\_GNN can capture the key subgraphs but include several additional edges. The results of the PGExplainer and GNNExplainer still contain several discrete edges. 

\begin{figure*}[ht!]
    \centering
    \includegraphics[width=2\columnwidth]{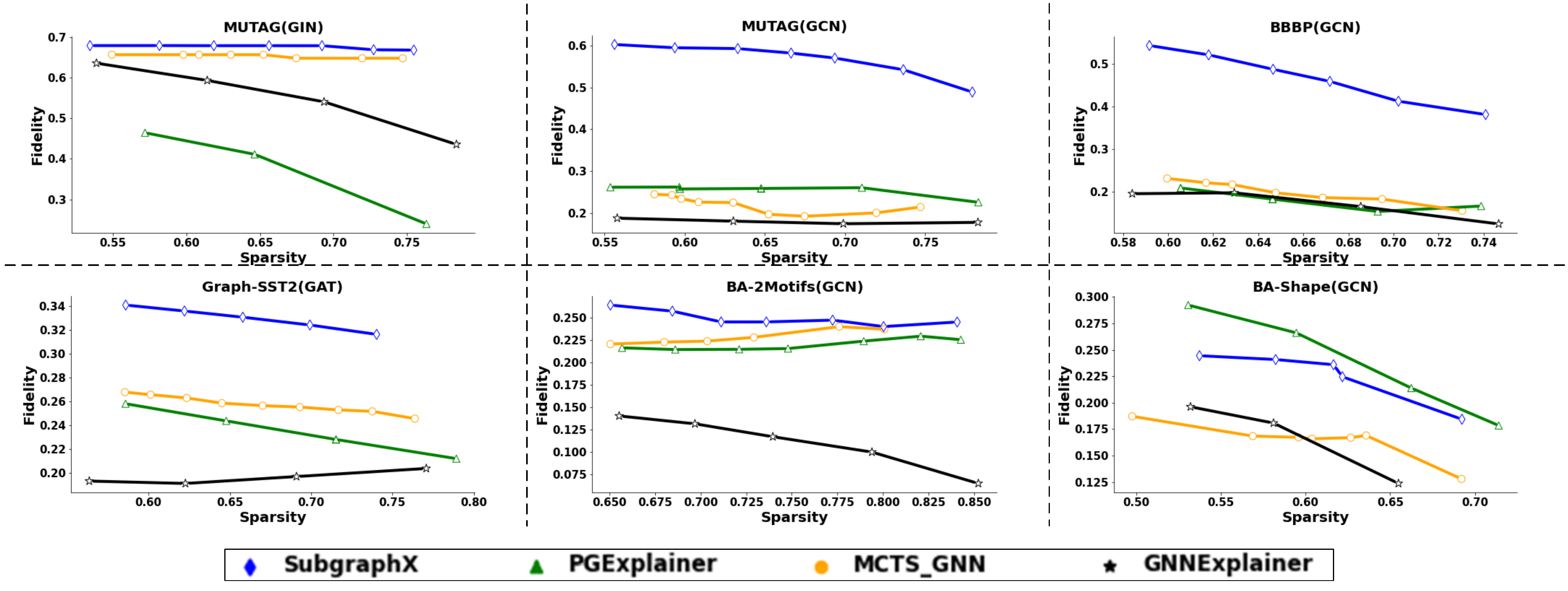}
    %\vspace{-0.4cm}
    \caption{The quantitative studies for different explanation methods. Note that since the Sparsity scores cannot be fully controlled, we compare different methods with Fidelity scores under similar similar levels of Sparsity.}
    \label{fig:vis5}
\end{figure*}

For the dataset Graph-SST2, we employ GATs as the graph model and report the results in Figure~\ref{fig:vis3}.
In the first row, the prediction is correct and the label is positive. Both our SubgraphX and the MCTS\_GNN can find word phrases with positive semantic meaning, such as ``makes old story new'', which can reasonably explain the prediction. The explanations provided by PGExplainer and GNNExplainer are, however, less semantically related. In the second row, the input is negative but the prediction is positive. All methods except PGExplainer can explain the decision that the GNN model regards positive phrases ``truly going to inspire'' as important, thus yielding a positive but incorrect prediction. It is noteworthy that our method tends to include fewer neural words, such as ``the'', ``me'', and ``screen'', etc.

Overall, our SubgraphX can explain both correct and incorrect predictions for different graph data and GNN models. Our explanations are more 
human-intelligible than comparing methods. 
More results for graph classification models are reported in Supplementary Section~\ref{sup:gc}.

\subsection{Explanations for Node Classification Models}
\begin{figure}[t]
    \centering
    \includegraphics[width=\columnwidth]{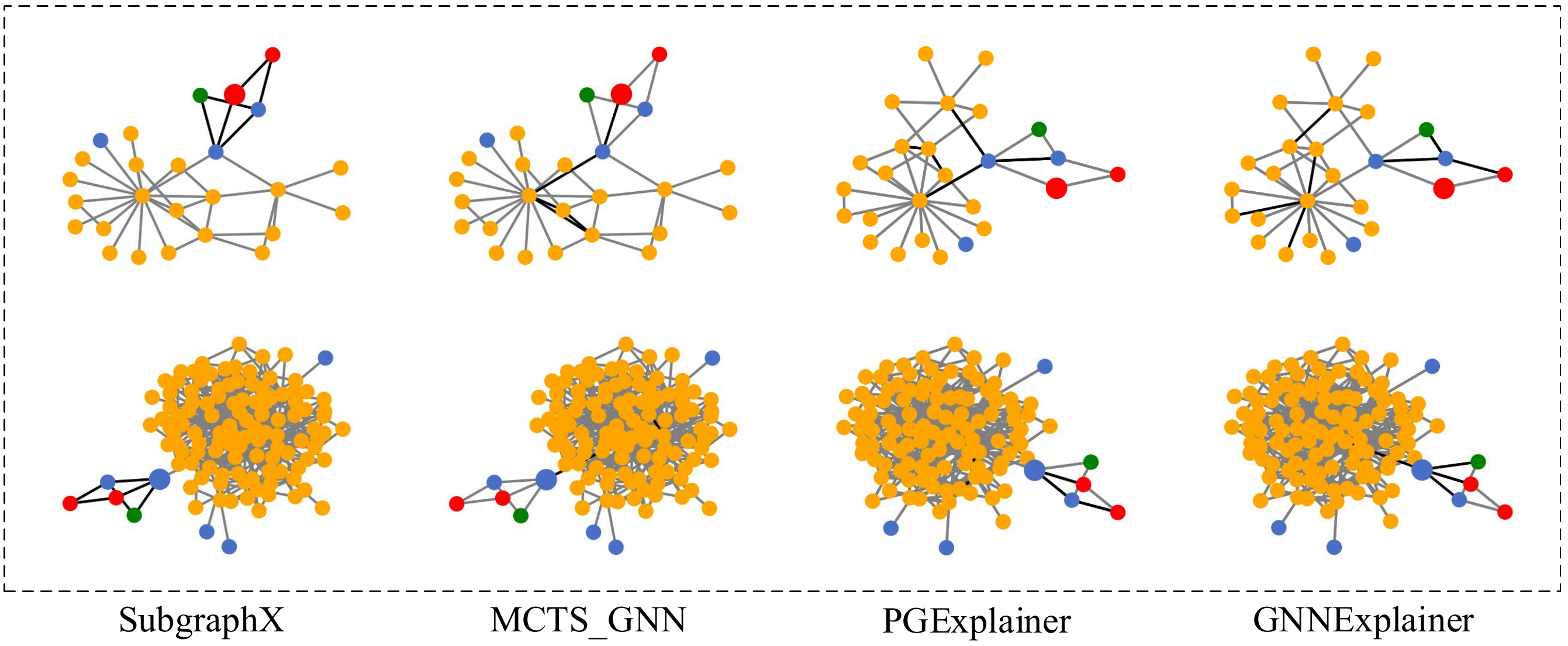}
    %\vspace{-0.7cm}
    \caption{Explanation results on the BA-Shape dataset. The target node is shown in a larger size. Different colors denote node labels.}
    \label{fig:vis4}
\end{figure}

We also compare different methods on the node classification tasks. We use the BA-Shape dataset and train a GCN model to perform node classification.  The visualization results are reported in Figure~\ref{fig:vis4} where the important substructures are shown in bold. We can verify if the explanations are consistent with the rules (the motifs) to label different nodes. 
For both examples, the target nodes are correctly classified.
Obviously, our SubgraphX is precisely targeting the motifs as the explanations, which is reasonable and promising. For other methods, their explanations only cover partial motifs and include other structures.  
More results are reported in Supplementary Section~\ref{sup:nc}. 

\begin{table*}
    \caption{Efficiency studies of different methods.
    }
    \label{Analysis}
    \centering
    \begin{tabular}{@{}lccccc@{}}
        \toprule[1.5pt]
        \textbf{Method}  & \textbf{MCTS$^*$}     & \textbf{MCTS$^\dagger$}& \textbf{SubgraphX} & \textbf{GNNExplainer} &  \textbf{PGExplainer}    \\
        \midrule[1pt] 
        \textsc{Time}  & $>$10 hours  & $865.4\pm1.6$s  & $77.8\pm3.8$s & $16.2\pm0.2$s & $0.02$s (Training 362s)   \\
        \textsc{Fidelity} & N/A & 0.53  & 0.55 & 0.19 &  0.18 \\
        \bottomrule[1pt]
    \end{tabular}
\end{table*}
\subsection{Quantitative Studies}
While visualizations are important to evaluate different explanation methods, human evaluations may not be accurate due to the lack of ground truths. Hence, we further conduct quantitative studies to compare these methods. Specifically, we employ the metrics Fidelity and Sparsity to evaluate explanation results~\cite{pope2019explainability, yuan2020explainability}. The Fidelity metric measures whether the explanations are faithfully important to the model's predictions. It removes the important structures from the input graphs and computes the difference between predictions. In addition, the Sparsity metric measures the fraction of structures that are identified as important by explanation methods. Note that high Sparsity scores mean smaller structures are identified as important, which can affect the Fidelity scores since smaller structures (high Sparsity) tend to be less important (low Fidelity). Hence, for fair comparisons, we compare different methods using Fidelity under similar levels of Sparsity. The results are reported in Figure~\ref{fig:vis5} where we plot the curves of Fidelity scores with respect to the Sparsity scores. Obviously, for five out of six experiments, our proposed method outperforms the comparing methods significantly and consistently under different sparsity levels. For the BA-Shape (GCN) experiment, our SubgraphX obtains slightly lower but still competitive  Fidelity scores compared with the PGExplainer. Overall, such results indicate that the explanations of our method are more faithful and important to the GNN models.
More details of evaluation metrics are introduced in Supplementary Section~\ref{sup:data}.

\subsection{Efficiency Studies} \label{article:efficiency}
Finally, we study the efficiency of our proposed method. For 50 graphs with an average of 24.96 nodes from the BBBP dataset, we show the averaging time cost to obtain explanations for each graph. We repeat the experiments 3 times and report the results Table~\ref{Analysis}.  Here MCTS$^*$ denotes the baseline that follows Eq. (\ref{eq:9}) to compute Shapley values. Compared with our SubgraphX, the difference is the usage of Monte Carlo sampling. In addition, MCTS$^\dagger$ indicates the baseline computing Shapley values with Monte Carlo sampling but without our proposed approximation schemes. Specifically, MCTS$^\dagger$ samples coalition sets from the player set $P$ instead of the reduced set $P'$.
First, the time cost of MCTS$^*$ is extremely high since it needs to enumerate all possible coalition sets. Next, compared with MCTS$^\dagger$, our SubgraphX is 11 times faster while the obtained explanations have similar Fidelity scores. It demonstrates our approximation schemes are both effective and efficient. Even though our method is slower than GNNExplainer and PGExplainer, the Fidelity scores of our explanations are 300\% higher than theirs. 
{Furthermore, 
the PGExplainer requires to train its model using the whole dataset. While 
offline model training cost is not directly comparable, the additional and significant time cost may be an issue when the dataset is large-scale.} Considering our explanations are with higher-quality and more human-intelligible, we believe such time complexity is reasonable and acceptable. 

\subsection{Study of Pruning Actions}\label{sup:pa}
Finally, we discuss the pruning actions in our MCTS. For 
the graph associated with each non-leaf tree search node, we perform node pruning to obtain its children subgraphs. Specifically, when a node is removed, all edges connected with it are also removed. In addition, if multiple disconnected subgraphs are obtained after removing a node, only the largest subgraph is kept. Instead of exploring all possible node pruning actions, we explore two strategies: Low2high and High2low. First, Low2high arranges the nodes based on their node degrees from low to high and only considers the pruning actions corresponding to the first $k$ low degree nodes. Meanwhile, High2low arranges the nodes in order from high degree to low degree and only considers the first $k$ high degree nodes for pruning. Intuitively, High2low is more efficient but may ignore the optimal solutions. In this work, we employ the High2low strategy for BA-Shape(GCNs), and Low2high strategy for other models, and set the k to 12 for all the datasets.
\begin{table}
    \caption{The studies of different pruning strategies. 
    }
    \label{table:prun}
    \centering
    \begin{tabular}{@{}lccc@{}}
        \toprule[1.5pt]
        \textbf{Method}  & Time     &  Fidelity    \\
        \midrule[1pt] 
        \textsc{Low2high}  & 107.24s  & 0.66149     \\
        \textsc{High2low} & 21.52s & 0.61046  \\
        \bottomrule[1pt]
    \end{tabular}
\end{table}
We conduct experiments to analyze these two pruning strategies for our SubgraphX algorithm and show the average time cost and Fidelity score in Table~\ref{table:prun}. Specifically, we randomly select 50 graphs from the BBBP datasets with an average node number of  24.96, which is the same in Section ~\ref{article:efficiency}. In addition, we
set Monte-Carlo sampling steps $T$ to 100, and select the subgraphs with the highest Shapley values and contain less than 15 nodes to calculate the Fidelity. Obviously, High2low is 5 times faster than Low2high but the Fidelity scores are inferior. 

\section{Conclusions} 
While considerable efforts have been devoted to study the explainability of GNNs, none of existing methods can explain GNN predictions with subgraphs. We argue that subgraphs are building blocks of complex graphs and are more human-intelligible. To this end, we propose the SubgraphX to explain GNNs by identifying important subgraphs explicitly. We employ the Monte Carlo tree search algorithm to efficiently explore different subgraph. For each subgraph, we propose to employ Shapley values to measure its importance by considering the interactions among different graph structures. To expedite computations, we propose efficient approximation schemes to compute Shapley values by considering interactions only within the information aggregation range. Experimental results show our SubgraphX obtain higher-quality and more human-intelligible explanations while keeping time complexity acceptable.

%\section*{Acknowledgements}
%This work was supported in part by National Science Foundation grants XXXX and XXXX.


\bibliography{xsub}

\begin{thebibliography}{49}
\providecommand{\natexlab}[1]{#1}
\providecommand{\url}[1]{\texttt{#1}}
\expandafter\ifx\csname urlstyle\endcsname\relax
  \providecommand{\doi}[1]{doi: #1}\else
  \providecommand{\doi}{doi: \begingroup \urlstyle{rm}\Url}\fi

\bibitem[Alon(2007)]{alon2007network}
Alon, U.
\newblock Network motifs: theory and experimental approaches.
\newblock \emph{Nature Reviews Genetics}, 8\penalty0 (6):\penalty0 450--461,
  2007.

\bibitem[Alon(2019)]{alon2019introduction}
Alon, U.
\newblock \emph{An introduction to systems biology: design principles of
  biological circuits}.
\newblock CRC press, 2019.

\bibitem[Baldassarre \& Azizpour(2019)Baldassarre and
  Azizpour]{baldassarre2019explainability}
Baldassarre, F. and Azizpour, H.
\newblock Explainability techniques for graph convolutional networks.
\newblock In \emph{International Conference on Machine Learning (ICML)
  Workshops, 2019 Workshop on Learning and Reasoning with Graph-Structured
  Representations}, 2019.

\bibitem[Chen et~al.(2018{\natexlab{a}})Chen, Song, Wainwright, and
  Jordan]{chen2018shapley}
Chen, J., Song, L., Wainwright, M.~J., and Jordan, M.~I.
\newblock L-shapley and c-shapley: Efficient model interpretation for
  structured data.
\newblock In \emph{International Conference on Learning Representations},
  2018{\natexlab{a}}.

\bibitem[Chen et~al.(2018{\natexlab{b}})Chen, Li, and
  Bruna]{chen2017supervised}
Chen, Z., Li, L., and Bruna, J.
\newblock Supervised community detection with line graph neural networks.
\newblock In \emph{International Conference on Learning Representations},
  2018{\natexlab{b}}.

\bibitem[Debnath et~al.(1991)Debnath, Lopez~de Compadre, Debnath, Shusterman,
  and Hansch]{debnath1991structure}
Debnath, A.~K., Lopez~de Compadre, R.~L., Debnath, G., Shusterman, A.~J., and
  Hansch, C.
\newblock Structure-activity relationship of mutagenic aromatic and
  heteroaromatic nitro compounds. correlation with molecular orbital energies
  and hydrophobicity.
\newblock \emph{Journal of medicinal chemistry}, 34\penalty0 (2):\penalty0
  786--797, 1991.

\bibitem[Devlin et~al.(2019)Devlin, Chang, Lee, and Toutanova]{devlin2018bert}
Devlin, J., Chang, M.-W., Lee, K., and Toutanova, K.
\newblock {BERT}: Pre-training of deep bidirectional transformers for language
  understanding.
\newblock In \emph{Proceedings of the 2019 Conference of the North {A}merican
  Chapter of the Association for Computational Linguistics: Human Language
  Technologies, Volume 1}, pp.\  4171--4186, Minneapolis, Minnesota, June 2019.
  Association for Computational Linguistics.
\newblock \doi{10.18653/v1/N19-1423}.
\newblock URL \url{https://www.aclweb.org/anthology/N19-1423}.

\bibitem[Du et~al.(2018)Du, Liu, Song, and Hu]{du2018towards}
Du, M., Liu, N., Song, Q., and Hu, X.
\newblock Towards explanation of dnn-based prediction with guided feature
  inversion.
\newblock In \emph{Proceedings of the 24th ACM SIGKDD International Conference
  on Knowledge Discovery \& Data Mining}, pp.\  1358--1367, 2018.

\bibitem[Gao \& Ji(2019)Gao and Ji]{gao2019graph}
Gao, H. and Ji, S.
\newblock Graph {U-Nets}.
\newblock In \emph{international conference on machine learning}, pp.\
  2083--2092. PMLR, 2019.

\bibitem[Gao et~al.(2018)Gao, Wang, and Ji]{Gao:KDD18}
Gao, H., Wang, Z., and Ji, S.
\newblock Large-scale learnable graph convolutional networks.
\newblock In \emph{Proceedings of the 24th ACM SIGKDD International Conference
  on Knowledge Discovery and Data Mining}, pp.\  1416--1424, 2018.

\bibitem[Gardner et~al.(2018)Gardner, Grus, Neumann, Tafjord, Dasigi, Liu,
  Peters, Schmitz, and Zettlemoyer]{gardner2018allennlp}
Gardner, M., Grus, J., Neumann, M., Tafjord, O., Dasigi, P., Liu, N.~F.,
  Peters, M., Schmitz, M., and Zettlemoyer, L.
\newblock {A}llen{NLP}: A deep semantic natural language processing platform.
\newblock In \emph{Proceedings of Workshop for {NLP} Open Source Software
  ({NLP}-{OSS})}, pp.\  1--6, Melbourne, Australia, July 2018. Association for
  Computational Linguistics.
\newblock \doi{10.18653/v1/W18-2501}.
\newblock URL \url{https://www.aclweb.org/anthology/W18-2501}.

\bibitem[Gilmer et~al.(2017)Gilmer, Schoenholz, Riley, Vinyals, and
  Dahl]{gilmer2017neural}
Gilmer, J., Schoenholz, S.~S., Riley, P.~F., Vinyals, O., and Dahl, G.~E.
\newblock Neural message passing for quantum chemistry.
\newblock In \emph{International Conference on Machine Learning}, pp.\
  1263--1272. PMLR, 2017.

\bibitem[Huang et~al.(2020)Huang, Yamada, Tian, Singh, Yin, and
  Chang]{huang2020graphlime}
Huang, Q., Yamada, M., Tian, Y., Singh, D., Yin, D., and Chang, Y.
\newblock Graphlime: Local interpretable model explanations for graph neural
  networks.
\newblock \emph{arXiv preprint arXiv:2001.06216}, 2020.

\bibitem[Jang et~al.(2016)Jang, Gu, and Poole]{jang2016categorical}
Jang, E., Gu, S., and Poole, B.
\newblock Categorical reparameterization with gumbel-softmax.
\newblock In \emph{International Conference on Learning Representations}, 2016.

\bibitem[Jin et~al.(2020)Jin, Barzilay, and Jaakkola]{jin2020multi}
Jin, W., Barzilay, R., and Jaakkola, T.
\newblock Multi-objective molecule generation using interpretable
  substructures.
\newblock In \emph{International Conference on Machine Learning}, pp.\
  4849--4859. PMLR, 2020.

\bibitem[Kipf \& Welling(2017)Kipf and Welling]{kipf2016semi}
Kipf, T.~N. and Welling, M.
\newblock Semi-supervised classification with graph convolutional networks.
\newblock In \emph{International Conference on Learning Representations}, 2017.

\bibitem[Kuhn \& Tucker(1953)Kuhn and Tucker]{kuhn1953contributions}
Kuhn, H.~W. and Tucker, A.~W.
\newblock \emph{Contributions to the Theory of Games}, volume~2.
\newblock Princeton University Press, 1953.

\bibitem[Liu et~al.(2021)Liu, Luo, Wang, Xie, Yuan, Gui, Yu, Xu, Zhang, Liu,
  Yan, Liu, Fu, Oztekin, Zhang, and Ji]{liu2021dig}
Liu, M., Luo, Y., Wang, L., Xie, Y., Yuan, H., Gui, S., Yu, H., Xu, Z., Zhang,
  J., Liu, Y., Yan, K., Liu, H., Fu, C., Oztekin, B., Zhang, X., and Ji, S.
\newblock {DIG}: A turnkey library for diving into graph deep learning
  research.
\newblock \emph{arXiv preprint arXiv:2103.12608}, 2021.

\bibitem[Lundberg \& Lee(2017)Lundberg and Lee]{lundberg2017unified}
Lundberg, S.~M. and Lee, S.-I.
\newblock A unified approach to interpreting model predictions.
\newblock In \emph{Advances in neural information processing systems}, pp.\
  4765--4774, 2017.

\bibitem[Luo et~al.(2020)Luo, Cheng, Xu, Yu, Zong, Chen, and
  Zhang]{luoparameterized}
Luo, D., Cheng, W., Xu, D., Yu, W., Zong, B., Chen, H., and Zhang, X.
\newblock Parameterized explainer for graph neural network.
\newblock In Larochelle, H., Ranzato, M., Hadsell, R., Balcan, M.~F., and Lin,
  H. (eds.), \emph{Advances in Neural Information Processing Systems},
  volume~33, pp.\  19620--19631. Curran Associates, Inc., 2020.

\bibitem[Milo et~al.(2002)Milo, Shen-Orr, Itzkovitz, Kashtan, Chklovskii, and
  Alon]{milo2002network}
Milo, R., Shen-Orr, S., Itzkovitz, S., Kashtan, N., Chklovskii, D., and Alon,
  U.
\newblock Network motifs: simple building blocks of complex networks.
\newblock \emph{Science}, 298\penalty0 (5594):\penalty0 824--827, 2002.

\bibitem[Pope et~al.(2019)Pope, Kolouri, Rostami, Martin, and
  Hoffmann]{pope2019explainability}
Pope, P.~E., Kolouri, S., Rostami, M., Martin, C.~E., and Hoffmann, H.
\newblock Explainability methods for graph convolutional neural networks.
\newblock In \emph{Proceedings of the IEEE Conference on Computer Vision and
  Pattern Recognition}, pp.\  10772--10781, 2019.

\bibitem[Schlichtkrull et~al.(2021{\natexlab{a}})Schlichtkrull, Cao, and
  Titov]{schlichtkrull2020interpreting}
Schlichtkrull, M.~S., Cao, N.~D., and Titov, I.
\newblock Interpreting graph neural networks for {\{}nlp{\}} with
  differentiable edge masking.
\newblock In \emph{International Conference on Learning Representations},
  2021{\natexlab{a}}.
\newblock URL \url{https://openreview.net/forum?id=WznmQa42ZAx}.

\bibitem[Schlichtkrull et~al.(2021{\natexlab{b}})Schlichtkrull, Cao, and
  Titov]{schlichtkrull2021interpreting}
Schlichtkrull, M.~S., Cao, N.~D., and Titov, I.
\newblock Interpreting graph neural networks for {\{}nlp{\}} with
  differentiable edge masking.
\newblock In \emph{International Conference on Learning Representations},
  2021{\natexlab{b}}.
\newblock URL \url{https://openreview.net/forum?id=WznmQa42ZAx}.

\bibitem[Schnake et~al.(2020)Schnake, Eberle, Lederer, Nakajima, Sch{\"u}tt,
  M{\"u}ller, and Montavon]{schnake2020xai}
Schnake, T., Eberle, O., Lederer, J., Nakajima, S., Sch{\"u}tt, K.~T.,
  M{\"u}ller, K.-R., and Montavon, G.
\newblock Xai for graphs: Explaining graph neural network predictions by
  identifying relevant walks.
\newblock \emph{arXiv preprint arXiv:2006.03589}, 2020.

\bibitem[Schwarzenberg et~al.(2019)Schwarzenberg, H{\"u}bner, Harbecke, Alt,
  and Hennig]{schwarzenberg2019layerwise}
Schwarzenberg, R., H{\"u}bner, M., Harbecke, D., Alt, C., and Hennig, L.
\newblock Layerwise relevance visualization in convolutional text graph
  classifiers.
\newblock \emph{arXiv preprint arXiv:1909.10911}, 2019.

\bibitem[Shen-Orr et~al.(2002)Shen-Orr, Milo, Mangan, and
  Alon]{shen2002network}
Shen-Orr, S.~S., Milo, R., Mangan, S., and Alon, U.
\newblock Network motifs in the transcriptional regulation network of
  escherichia coli.
\newblock \emph{Nature genetics}, 31\penalty0 (1):\penalty0 64--68, 2002.

\bibitem[Silver et~al.(2017)Silver, Schrittwieser, Simonyan, Antonoglou, Huang,
  Guez, Hubert, Baker, Lai, Bolton, et~al.]{silver2017mastering}
Silver, D., Schrittwieser, J., Simonyan, K., Antonoglou, I., Huang, A., Guez,
  A., Hubert, T., Baker, L., Lai, M., Bolton, A., et~al.
\newblock Mastering the game of go without human knowledge.
\newblock \emph{nature}, 550\penalty0 (7676):\penalty0 354--359, 2017.

\bibitem[Simonyan et~al.(2013)Simonyan, Vedaldi, and
  Zisserman]{simonyan2013deep}
Simonyan, K., Vedaldi, A., and Zisserman, A.
\newblock Deep inside convolutional networks: Visualising image classification
  models and saliency maps.
\newblock \emph{arXiv preprint arXiv:1312.6034}, 2013.

\bibitem[Smilkov et~al.(2017)Smilkov, Thorat, Kim, Vi{\'e}gas, and
  Wattenberg]{smilkov2017smoothgrad}
Smilkov, D., Thorat, N., Kim, B., Vi{\'e}gas, F., and Wattenberg, M.
\newblock Smoothgrad: removing noise by adding noise.
\newblock \emph{arXiv preprint arXiv:1706.03825}, 2017.

\bibitem[Springenberg et~al.(2015)Springenberg, Dosovitskiy, Brox, and
  Riedmiller]{springenberg2014striving}
Springenberg, J.~T., Dosovitskiy, A., Brox, T., and Riedmiller, M.
\newblock Striving for simplicity: The all convolutional net.
\newblock \emph{International Conference on Learning Representations}, 2015.

\bibitem[{\v{S}}trumbelj \& Kononenko(2014){\v{S}}trumbelj and
  Kononenko]{vstrumbelj2014explaining}
{\v{S}}trumbelj, E. and Kononenko, I.
\newblock Explaining prediction models and individual predictions with feature
  contributions.
\newblock \emph{Knowledge and information systems}, 41\penalty0 (3):\penalty0
  647--665, 2014.

\bibitem[Veličković et~al.(2018)Veličković, Cucurull, Casanova, Romero,
  Liò, and Bengio]{velivckovic2017graph}
Veličković, P., Cucurull, G., Casanova, A., Romero, A., Liò, P., and Bengio,
  Y.
\newblock Graph attention networks.
\newblock In \emph{International Conference on Learning Representations}, 2018.

\bibitem[Vu \& Thai(2020)Vu and Thai]{vu2020pgm}
Vu, M.~N. and Thai, M.~T.
\newblock Pgm-explainer: Probabilistic graphical model explanations for graph
  neural networks.
\newblock In \emph{Advances in neural information processing systems}, 2020.

\bibitem[Wang et~al.(2019)Wang, Ji, Shi, Wang, Ye, Cui, and
  Yu]{wang2019heterogeneous}
Wang, X., Ji, H., Shi, C., Wang, B., Ye, Y., Cui, P., and Yu, P.~S.
\newblock Heterogeneous graph attention network.
\newblock In \emph{The World Wide Web Conference}, pp.\  2022--2032, 2019.

\bibitem[Wang et~al.(2020)Wang, Liu, Luo, Xu, Xie, Wang, Cai, and
  Ji]{wang2020advanced}
Wang, Z., Liu, M., Luo, Y., Xu, Z., Xie, Y., Wang, L., Cai, L., and Ji, S.
\newblock Advanced graph and sequence neural networks for molecular property
  prediction and drug discovery.
\newblock \emph{arXiv preprint arXiv:2012.01981}, 2020.

\bibitem[Wu et~al.(2018)Wu, Ramsundar, Feinberg, Gomes, Geniesse, Pappu,
  Leswing, and Pande]{wu2018moleculenet}
Wu, Z., Ramsundar, B., Feinberg, E.~N., Gomes, J., Geniesse, C., Pappu, A.~S.,
  Leswing, K., and Pande, V.
\newblock Moleculenet: a benchmark for molecular machine learning.
\newblock \emph{Chemical science}, 9\penalty0 (2):\penalty0 513--530, 2018.

\bibitem[Xu et~al.(2019)Xu, Hu, Leskovec, and Jegelka]{xu2018powerful}
Xu, K., Hu, W., Leskovec, J., and Jegelka, S.
\newblock How powerful are graph neural networks?
\newblock In \emph{International Conference on Learning Representations}, 2019.
\newblock URL \url{https://openreview.net/forum?id=ryGs6iA5Km}.

\bibitem[Yang et~al.(2019)Yang, Pentyala, Mohseni, Du, Yuan, Linder, Ragan, Ji,
  and Hu]{yang2019xfake}
Yang, F., Pentyala, S.~K., Mohseni, S., Du, M., Yuan, H., Linder, R., Ragan,
  E.~D., Ji, S., and Hu, X.
\newblock Xfake: explainable fake news detector with visualizations.
\newblock In \emph{The World Wide Web Conference}, pp.\  3600--3604, 2019.

\bibitem[Ying et~al.(2019)Ying, Bourgeois, You, Zitnik, and
  Leskovec]{ying2019gnnexplainer}
Ying, Z., Bourgeois, D., You, J., Zitnik, M., and Leskovec, J.
\newblock Gnnexplainer: Generating explanations for graph neural networks.
\newblock In \emph{Advances in neural information processing systems}, pp.\
  9244--9255, 2019.

\bibitem[Yuan \& Ji(2020)Yuan and Ji]{Yuan2020StructPool}
Yuan, H. and Ji, S.
\newblock Structpool: Structured graph pooling via conditional random fields.
\newblock In \emph{International Conference on Learning Representations}, 2020.

\bibitem[Yuan \& Ji(2021)Yuan and Ji]{yuan2021node2seq}
Yuan, H. and Ji, S.
\newblock Node2seq: Towards trainable convolutions in graph neural networks.
\newblock \emph{arXiv preprint arXiv:2101.01849}, 2021.

\bibitem[Yuan et~al.(2019)Yuan, Chen, Hu, and Ji]{yuan2019Interpreting}
Yuan, H., Chen, Y., Hu, X., and Ji, S.
\newblock Interpreting deep models for text analysis via optimization and
  regularization methods.
\newblock In \emph{AAAI-19: Thirty-Third AAAI Conference on Artificial
  Intelligence}. Association for the Advancement of Artificial Intelligence,
  2019.

\bibitem[Yuan et~al.(2020{\natexlab{a}})Yuan, Cai, Hu, Wang, and
  Ji]{yuan2020interpreting}
Yuan, H., Cai, L., Hu, X., Wang, J., and Ji, S.
\newblock Interpreting image classifiers by generating discrete masks.
\newblock \emph{IEEE Transactions on Pattern Analysis and Machine
  Intelligence}, 2020{\natexlab{a}}.

\bibitem[Yuan et~al.(2020{\natexlab{b}})Yuan, Tang, Hu, and Ji]{xgnn}
Yuan, H., Tang, J., Hu, X., and Ji, S.
\newblock {XGNN}: Towards model-level explanations of graph neural networks.
\newblock KDD '20, pp.\  430–438, New York, NY, USA, 2020{\natexlab{b}}.
  Association for Computing Machinery.
\newblock ISBN 9781450379984.
\newblock \doi{10.1145/3394486.3403085}.
\newblock URL \url{https://doi.org/10.1145/3394486.3403085}.

\bibitem[Yuan et~al.(2020{\natexlab{c}})Yuan, Yu, Gui, and
  Ji]{yuan2020explainability}
Yuan, H., Yu, H., Gui, S., and Ji, S.
\newblock Explainability in graph neural networks: A taxonomic survey.
\newblock \emph{arXiv preprint arXiv:2012.15445}, 2020{\natexlab{c}}.

\bibitem[Zeiler \& Fergus(2014)Zeiler and Fergus]{zeiler2014visualizing}
Zeiler, M.~D. and Fergus, R.
\newblock Visualizing and understanding convolutional networks.
\newblock In \emph{European conference on computer vision}, pp.\  818--833.
  Springer, 2014.

\bibitem[Zhang et~al.(2018)Zhang, Cui, Neumann, and Chen]{zhang2018end}
Zhang, M., Cui, Z., Neumann, M., and Chen, Y.
\newblock An end-to-end deep learning architecture for graph classification.
\newblock In \emph{AAAI}, volume~18, pp.\  4438--4445, 2018.

\bibitem[Zhou et~al.(2016)Zhou, Khosla, Lapedriza, Oliva, and
  Torralba]{zhou2016learning}
Zhou, B., Khosla, A., Lapedriza, A., Oliva, A., and Torralba, A.
\newblock Learning deep features for discriminative localization.
\newblock In \emph{Proceedings of the IEEE Conference on Computer Vision and
  Pattern Recognition}, pp.\  2921--2929, 2016.

\end{thebibliography}
\bibliographystyle{icml2021}


\clearpage

\onecolumn
\icmltitle{On Explainability of Graph Neural Networks via Subgraph Explorations: Appendix}
%\twocolumn

\appendix
%\section{Appendix}

\section{Datasets and Experimental Settings}\label{sup:data}

\subsection{Datasets and GNN Models}
We employ different GNN variants to fit these datasets and explain the trained GNNs. Note that these models are trained to obtain reasonable performance. 
Specifically, we report the architectures and performance of these GNNs as below:
\begin{itemize}[noitemsep, topsep=0pt,leftmargin=*]
\item \textbf{MUTAG (GCNs)}: This GNN model consists of 3 GCN layers. The input feature dimension is 7 and the output dimensions of different GCN layers are set to 128, 128, 128, respectively. We employ max-pooling as the readout function and ReLU as the activation function. The model is trained for 2000 epochs with a learning rate of 0.005 and the testing accuracy is 0.92. We study the explanations for the whole dataset.

\item \textbf{MUTAG (GINs)}: This GNN model consists of 3 GIN layers. For each GIN layer, the MLP for feature transformations is a two-layer MLP. The input feature dimension is 7 and the output dimensions of different GIN layers are set to 128, 128, 128 respectively. We employ max-pooling as the readout function and ReLU as the activation function.
The model is trained for 2000 epochs with a learning rate of 0.005 and the testing accuracy is 1.00. We study the explanations for the whole dataset.

\item \textbf{BBBP (GCNs)}: This GNN model consists of 3 GCN layers. The input feature dimension is 9 and the output dimensions of different GCN layers are set to 128, 128, 128, respectively. We employ max-pooling as the readout function and ReLU as the activation function.
The model is trained for 800 epochs with a learning rate of 0.005 and the testing accuracy is 0.863. We randomly split this dataset into the training set (80\%), validation set (10\%), and testing set (10\%). We study the explanations for the testing set.

\item \textbf{Graph-SST2 (GATs)}: This GNN model consists of 3 GAT layers. The input feature dimension is 768 and all GAT layers have 10 heads with 10-dimensional features. We employ max-pooling as the readout function and ReLU as the activation function. In addition, we set the dropout rate to 0.6 to avoid overfitting. 
The model is trained for 800 epochs with a learning rate of 0.005 and the testing accuracy is 0.881. We follow the training, validation, and testing splitting of the original SST2 dataset. 
We study the explanations for the testing set.

\item \textbf{BA-2Motifs (GCNs)}: This GNN model consists of 3 GCN layers. The input feature dimension is 10 and the output dimensions of different GCN layers are set to 20, 20, 20, respectively. For each GCN layer, we employ L2 normalization to normalize node features. We employ average pooling as the readout function and ReLU as the activation function. The model is trained for 800 epochs with a learning rate of 0.005 and the testing accuracy is 0.99. We randomly split this dataset into the training set (80\%), validation set (10\%), and testing set (10\%). We study the explanations for the testing set.

\item \textbf{BA-Shape (GCNs)}: This GNN model consists of 3 GCN layers. The input feature dimension is 10 and the output dimensions of different GCN layers are set to 20, 20, 20, respectively. For each GCN layer, we employ L2 normalization to normalize node features. In addition, we use ReLU as the activation function. 
The model is trained for 800 epochs with a learning rate of 0.005 and the testing accuracy is 0.957. We randomly split this dataset into the training set (80\%), validation set (10\%), and testing set (10\%). We study the explanations for the testing set.

\end{itemize}

\subsection{Experimental Settings}

\subsection{Evaluation Metrics}
We further introduce the evaluation metrics in detail. First, given a graph $G_i$, its prediction class $y_i$, and its explanation, we obtain a hard explanation mask $M_i$ where each element is 0 or 1 to indicate whether the corresponding node is identified as important. For our SubgraphX and MCTS-based baselines, the masks can be directly determined by the obtained subgraphs. For GNNExplainer and PGExplainer, their explanations are edge masks and can be converted to explanation masks by selecting the nodes connected with these important edges. Then by occluding the important nodes in $G_i$ based on $M_i$, we can obtain a new graph $\hat{G}_i$. Finally, the Fidelity score can be computed as 
\begin{equation}
    Fidelity = \frac{1}{N}\sum_{i=1}^{N} (f(G_i)_{y_i} - f(\hat{G}_i)_{y_i}),
\end{equation}
where $N$ is the total number of testing samples, $f(G_i)_{y_i}$ means the predicted probability of class $y_i$ for the original graph $G_i$. Intuitively, Fidelity measures the averaged probability change for the predictions by removing important input features. Since simply removing nodes significantly affect the graph structures, we occlude these nodes with zero features to compute the Fidelity. In addition, we also employ Sparsity to measure the fraction of nodes are selected in the explanations. Then it can be computed as 
\begin{equation}
    Sparsity = \frac{1}{N}\sum_{i=1}^{N}(1 - \frac{|M_i|}{|G_i|}),
\end{equation}
where $|M_i|$ denotes the number of important nodes identified in $M_i$ and $|G_i|$ means the number of nodes in $G_i$. Ideally, good explanations should select fewer nodes (high Sparsity) but lead to significant prediction drops (high Fidelity).

\begin{figure*}[ht!]
    \centering
    \includegraphics[height=0.96\textheight,width=0.9\columnwidth]{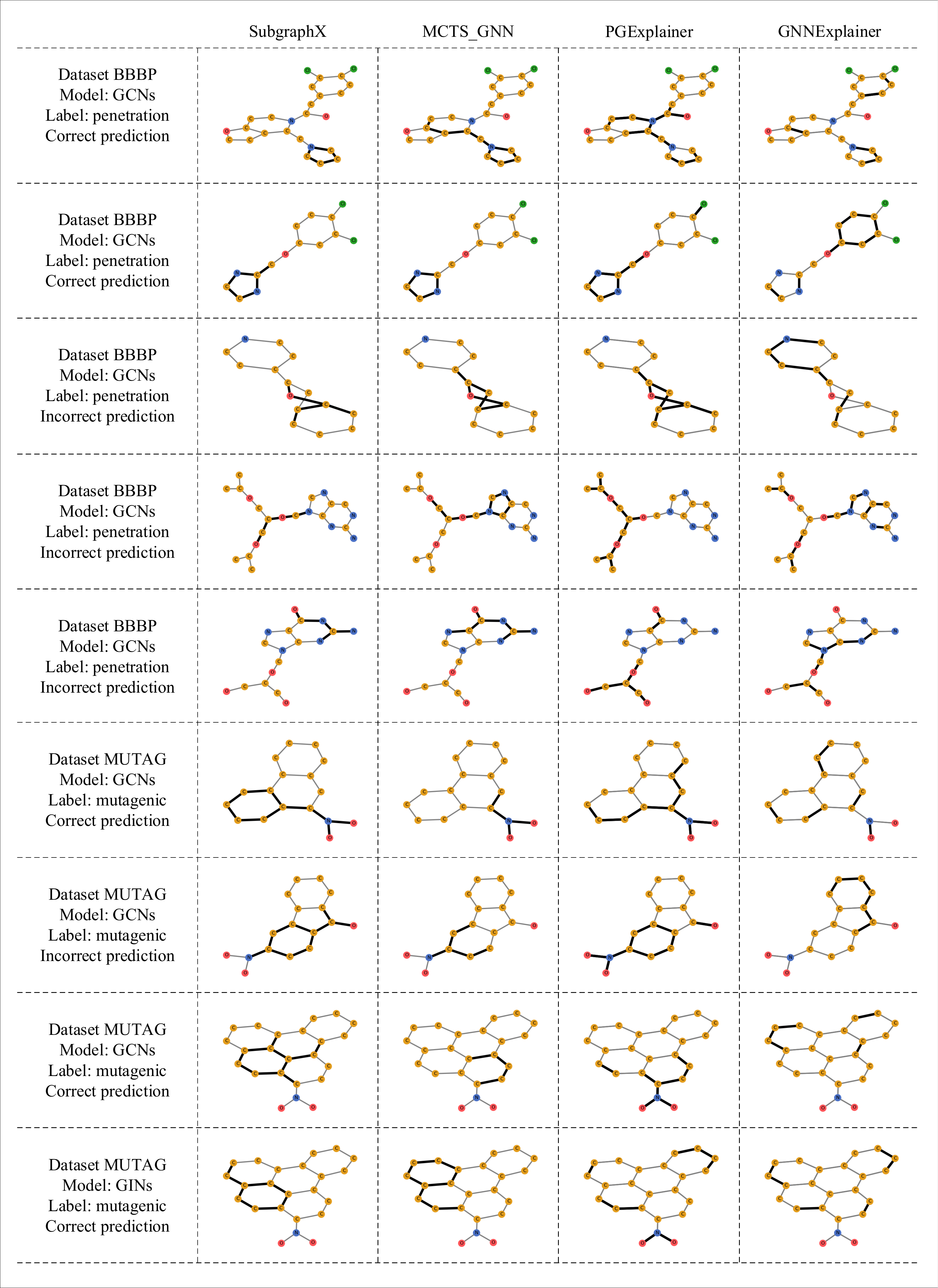}
    \caption{Explanation results of the BBBP and MUTAG datasets. Here Carbon, Oxygen, Nitrogen, and Chlorine are shown in yellow, red, and blue, green respectively.}
    \label{supp_vis1}
\end{figure*}

\begin{figure*}[ht!]
    \centering
    \includegraphics[height=0.75\textheight,width=\columnwidth]{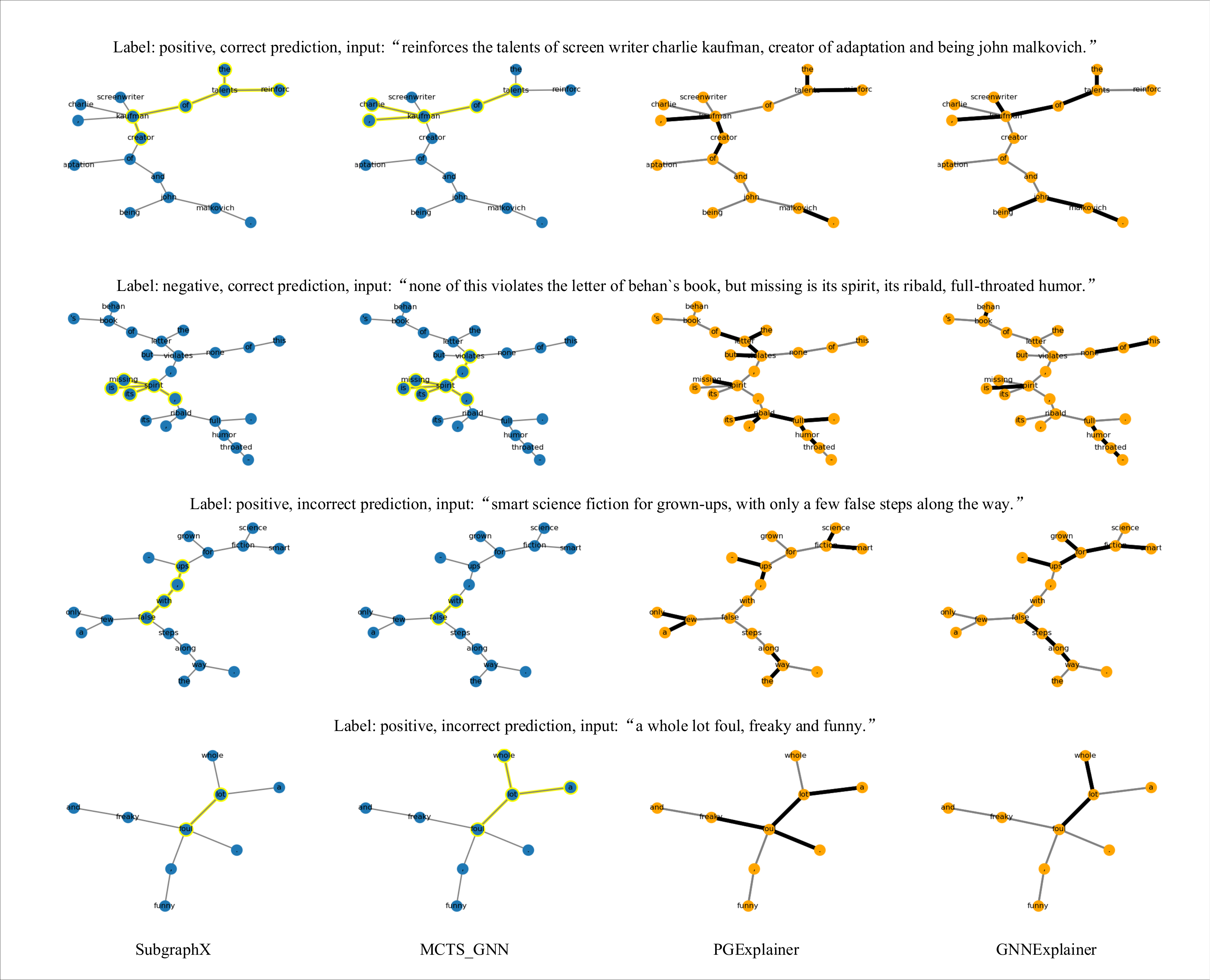}
    \caption{Explanation results of Grpah-SST2 dataset.}
    \label{supp_vis2}
\end{figure*}

\begin{figure*}[ht!]
    \centering
    \includegraphics[width=\columnwidth]{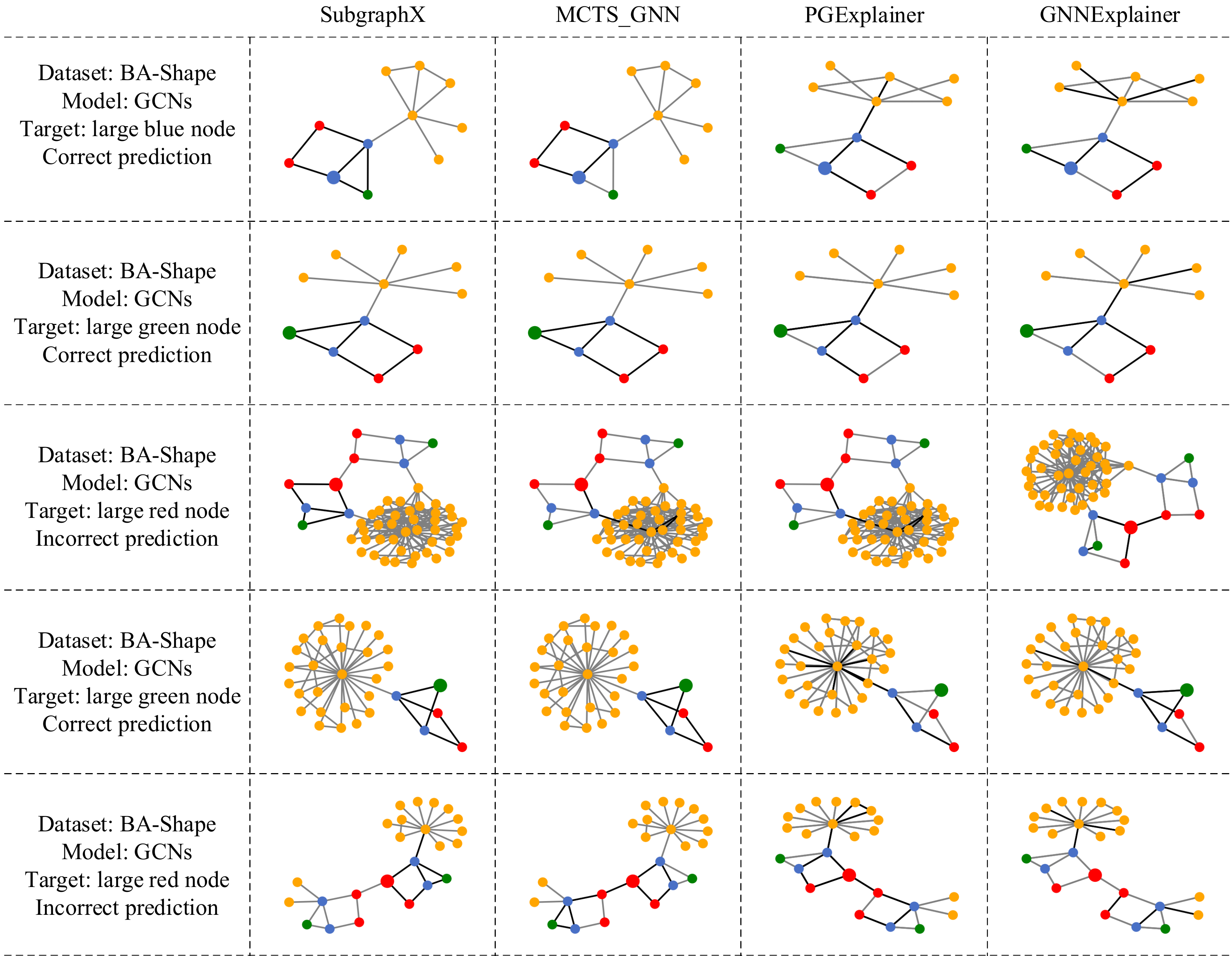}
    \caption{Explanation results of BA-Shape dataset. The target node is shown in a larger size.  }
    \label{supp_vis3}
\end{figure*}

\section{Explanations for Graph Classification Models}\label{sup:gc}
In this section, we report more visualizations of explanations for graph classification models. The results are reported in Figure~\ref{supp_vis1} and~\ref{supp_vis2}. In Figure~\ref{supp_vis1}, we show the explanations of real-world datasets BBBP and MUTAG. Obviously, our proposed method can provide more human-intelligible subgraphs as explanations while PGExplainer and GNNExplainer focus on discrete edges. In addition, we also report the results of sentiment dataset Graph-SST2 in Figure~\ref{supp_vis2}. The results show that our SubgraphX can provide reasonable explanations to explain the predictions. For example, in the second row, the input sentence is ``none of this violates the letter of behan`s book, but missing is its spirit, its ribald, full-throated humor'', whose label is negative and the prediction is correct. From the human's view, ``missing'' should be the keyword for the semantic meaning. Our SubgraphX shows that the ``missing is its spirit'' phrase is important, which successfully captures the keyword. The other methods capture the words and phrases such as ``violates'', ``none of this'', which are less related to the negative meaning.

\section{Explanations for Node Classification Models}\label{sup:nc}
In this section, we report more visualizations of explanations for node classification models. The results are reported in Figure~\ref{supp_vis3} where we show the explanations of node classification dataset BA-Shape. Obviously, our SubgprahX focuses on the whole motifs for correct predictions and captures partial motifs for incorrect predictions. This is reasonable since if the model can capture the whole motif, then it is expected to correctly predict the target node; otherwise, the information of partial motifs is not enough to make correct predictions.

\end{document}